\documentclass[journal]{IEEEtran}
\usepackage{amsmath}
\usepackage{amsfonts}
\usepackage{amssymb}
\usepackage{amsthm}
\usepackage{accents}
\usepackage{algorithmic}
\usepackage{soul}
\usepackage{xpatch}
\usepackage{array}
\usepackage[caption=false,font=normalsize,labelfont=sf,textfont=sf]{subfig}
\usepackage{textcomp}
\usepackage{stfloats}
\usepackage{multirow}
\usepackage{booktabs}
\usepackage{enumerate}
\usepackage{algorithm}
\usepackage{algorithmic}
\usepackage{url}
\usepackage{verbatim}
\usepackage{graphicx}
\makeatletter
\let\NAT@parse\undefined
\makeatother
\usepackage{hyperref} 
\hypersetup{
    colorlinks=true,
    linkcolor=blue,
    citecolor=blue,
    urlcolor=blue
}
\renewcommand{\algorithmicrequire}{\textbf{Input:}}
\renewcommand{\algorithmicensure}{\textbf{Output:}}
\def\BibTeX{{\rm B\kern-.05em{\sc i\kern-.025em b}\kern-.08em
    T\kern-.1667em\lower.7ex\hbox{E}\kern-.125emX}}
\usepackage{balance}
\begin{document}
\title{MGL2Rank: Learning to Rank the Importance of Nodes in Road Networks Based on Multi-Graph Fusion}
\author{Ming Xu, \IEEEmembership{Member, IEEE}, Jing Zhang
\vspace{-0.5cm}
\thanks{This work was supported in part by the Doctoral Scientific Research Foundation of Liaoning Technical University.

Corresponding author: Ming Xu.

M. Xu is with the software college, Liaoning Technical University. (\url{xum.2016@tsinghua.org.cn)}.

J. Zhang is with the software college, Liaoning Technical University. (\url{472121726@stu.lntu.edu.cn}).
}}
\maketitle
\begin{abstract}
The identification of important nodes with strong propagation capabilities in road networks is a vital topic in urban planning. Existing methods for evaluating the importance of nodes in traffic networks only consider topological information and traffic volumes, the diversity of the traffic characteristics in road networks, such as the number of lanes and average speed of road segments, is ignored, thus limiting their performance. To solve this problem, we propose a graph learning-based framework (MGL2Rank) that integrates the rich characteristics of road networks to rank the importance of nodes. This framework comprises an embedding module containing
a sampling algorithm (MGWalk) and an encoder network to learn the latent representations for
each road segment. MGWalk utilizes multi-graph fusion to capture the topology of road networks and establish associations between road segments based on their attributes. The obtained node representation is then used to learn the importance ranking of the road segments. Finally, a
synthetic dataset is constructed for ranking tasks based on the regional road network of Shenyang
City, and the ranking results on this dataset demonstrate the effectiveness of our method. The data and source code for MGL2Rank are available at \url{https://github.com/iCityLab/MGL2Rank}.
\end{abstract}

\section{Introduction}
\IEEEPARstart{U}{rban} road networks are highly complex and include numerous closely connected road segments and intersections. When certain important nodes are affected by traffic accidents, natural disasters, or other anomalies, they can cause traffic congestion in the surrounding areas and even large-scale collapse of road networks. Thus, the accurate identification of important nodes in a road network is crucial for assisting urban planning and improving network reliability and resistance to adversity.

Several studies have been conducted to identify important nodes in various networks. The methods proposed in these studies can be classified into three main categories: complex network theory-based methods, traditional
machine learning-based methods, and deep learning-based methods. For complex network theory-based methods, some studies \cite{freeman2002centrality, freeman1977set, 8392785} have evaluated the importance of nodes using different centrality metrics in general networks. In  \cite{agryzkov2012algorithm}, centrality metrics and node characteristics were combined to identify important nodes in an urban road network. Various algorithms \cite{xu2018discovery,yu2023ranking,li2020congestion,tsitsokas2023two,tempelmeier2021mining,wang2016feature,luo2021let} have been proposed based on different data sources, such as vehicle trajectory, traffic distribution, human movement, and social media data, to evaluate the importance of traffic nodes. However, these methods may not be sufficiently efficient and accurate for node ranking in traffic networks due to the lack of a correction mechanism. Traditional machine learning-based methods utilize various techniques, such as support vector regression machines with radial basis function (RBF) kernels \cite{ASGHARIANREZAEI2023119086} and gradient boosting decision tree (GBDT) \cite{DENG2019105652}, to identify important nodes. However, these methods are susceptible to loss of valuable information because they strongly depend on feature engineering. Deep learning-based methods have also been employed to rank the importance of nodes in diverse networks. For instance, graph convolutional neural networks and predicate-aware attention mechanisms have been used to identify important nodes in general networks \cite{yu2020identifying, park2019estimating}. The authors in 
\cite{huang2022traffic} clustered node embeddings based on topology and considered traffic volumes to evaluate the importance of intersections. However, these methods have limitations when applied to traffic networks. Specifically, they ignore the numerous characteristics of traffic networks that are important for node ranking, such as the capacity and speed limits of road segments. Generally, high-capacity road segments with high-speed limits should have higher rankings. Road segments with similar characteristics may have similar rankings because they tend to perform similar transportation functions. In summary, the shortcomings in identifying and ranking the importance of nodes in a traffic network are as follows.
\begin{enumerate}[(1)]
\itemsep=0pt
\item Limited attempts have been made to develop learnable methods for ranking the importance of nodes in traffic networks.
\item The influence of multiple roads and traffic characteristics on node importance in road networks is often overlooked.
\end{enumerate} 

In this paper, we propose a graph learning-based node importance ranking framework called MGL2Rank. This framework utilizes the topology and multiple characteristics of a road network to capture differences in the importance between road segments. Specifically, we first model the road network and traffic data in the form of multiple graphs and propose a sampling algorithm called MGWalk. This algorithm aggregates similar nodes by considering both the relationships of node adjacency and associations of attribute values. A representation learning module that leverages the output of MGWalk is developed to learn the embeddings of road segments. Subsequently, we introduce a learn-to-rank method to solve the pairwise ranking problem. Finally, a synthetic dataset of a real-world road network is constructed, and experiments are performed to demonstrate the effectiveness of the MGWalk algorithm and  MGL2Rank framework. The main contributions of this study are as follows:
\begin{itemize}
\item An embedding module containing a sampling algorithm (MGWalk) and an encoder network is proposed to learn the embeddings of road segments.
\item A ranking module that leverages the obtained embeddings to learn the importance ranking of the road segments is developed.
\item The effectiveness of the proposed method is verified by conducting experiments on the constructed dataset.
\end{itemize}

The remainder of this paper is organized as follows. In Section \ref{section.2}, we discuss related studies. In Section \ref{section.3}, we provide details of the proposed framework (MGL2Rank), including the MGWalk algorithm, embedding network, and learn-to-rank module. Section \ref{section.4} presents the experimental design and experimental results. In Section \ref{section.5}, we conclude the paper and discuss scope for future research.

\section{Related works}\label{section.2}
This study is closely related to three major recent research topics: (1) Evaluation of node importance, (2) Learning to rank, and (3) Graph representation learning.

\subsection{Evaluation of node importance}\vspace{0.3cm}
Existing methods employed to evaluate the importance of nodes in networks can be divided into four categories: (1) Node centrality analysis, (2) Link analysis, (3) Traditional machine learning, and (4) Deep learning.

Node centrality-based methods focus on the topology of a network with degree centrality \cite{freeman2002centrality} and betweenness centrality \cite{freeman1977set}, which are commonly used metrics. The degree centrality measures the local connectivity of a node, whereas the betweenness centrality captures its global impact on the network. In \cite{9537594}, fuzzy logic was employed to address the uncertainty in the important contributions of the neighboring nodes to the central node. Although centrality measures are computationally efficient, their sole reliance on the topology limits their applicability to complex tasks and large-scale networks.

Link analysis-based methods consider the link relationships and relative importance of nodes. For example, PageRank \cite{Page1999ThePC} analyzes the connections between nodes and iteratively calculates the importance of each node. HITS \cite{kleinberg1999authoritative} evaluates the influence of each node by iteratively calculating the hub and authority values. A survey \cite{2021A} showed the latest extensions and improvements to these methods. For traffic networks, the study \cite{agryzkov2012algorithm} utilized the PageRank vector to classify nodes based on their characteristics and generate various rankings of the node importance. Li et al. \cite{li2020congestion} constructed congestion propagation graphs and maximum spanning trees to identify bottlenecks in road networks. In another study \cite{luo2021let}, it was demonstrated that this problem is NP-hard, and a trajectory-driven traffic bottleneck identification framework was established. In our previous work \cite{xu2018discovery}, we proposed a data-driven framework to mine the node importance from comprehensive vehicle trajectory data. However, these methods are not learnable and cannot consider multiple traffic characteristics. Therefore, there is scope to improve their performance.

Traditional machine learning technologies have been applied to identify important nodes in various networks. In \cite{ASGHARIANREZAEI2023119086}, several machine learning models were trained on node representations to rank vital nodes. Zhao et al. \cite{9050733} captured the relationship between the real propagation ability of a node and its structural features by comparing different machine learning algorithms. Deng et al. \cite{DENG2019105652} identified insider trading nodes using a GBDT and differential evolution. However, these methods are inefficient and susceptible to information loss due to their reliance on feature engineering and domain knowledge.

Deep learning has been used to evaluate the importance of nodes. For general networks, the problem of identifying important nodes was transformed into a regression task using a graph convolutional neural network \cite{yu2020identifying}. In \cite{liu2022learning}, multi-task learning with a self-supervised pretext task was developed to extract information regarding the node location and global topology. GENI \cite{park2019estimating} employs a predicate-aware attention mechanism and flexible centrality adjustment for importance score aggregation in a knowledge graph. Few studies have used deep learning to identify the importance of nodes in traffic networks. In a representative study \cite{huang2022traffic}, a method called TraNode2vec was proposed. It clusters node embeddings based on topology and combines them with traffic volumes at intersections to evaluate the importance of intersections. However, the ranking results of TraNode2Vec were not determined by a learnable module, and the diverse characteristics of road networks could not be incorporated into the model. To solve these problems, we introduce a deep learning-based method that integrates multiple roads and traffic characteristics to identify and rank the importance of road segments. Unlike our previous method \cite{xu2018discovery}, the proposed method is learnable.

\subsection{Learning to rank}\vspace{0.3cm}
Learning to rank (LTR) \cite{liu2009learning} is a technique used to rank documents, items, and queries according to their relevance. This technique originated in the field of information retrieval and has been widely adopted across various domains, such as recommendation systems \cite{abdollahpouri2017controlling}, object detection \cite{tan2019learning}, and sentiment analysis \cite{song2017stock}.

Based on different input and output spaces, types of tasks, and objectives, LTR methods can be classified into three categories: pointwise, pairwise, and listwise methods. The pointwise method \cite{li2007mcrank} treats each object as an independent entity and transforms the ranking problem into a regression task. However, this method cannot account for the relative relationships between objects. The pairwise method \cite{burges2005learning} transforms the ranking problem into a binary classification task by combining objects into pairs and assigning labels based on their relevance. The objective of the model is to maximize the number of correct rankings. The study in \cite{maurya2021graph} achieved pairwise ranking using a graph neural network to approximate the betweenness and closeness centrality. Although the pairwise method is more accurate than the pointwise method, it ignores the overall positions of the objects in the ranking list. The listwise method \cite{xia2008listwise} is the most sophisticated LTR technique. In this method, the ranking problem is transformed into an optimization problem by treating the ranked list as a whole. The objective of the model is to maximize the relevance of all the objects in the candidate list. However, the listwise method is computationally expensive and highly sensitive to the data distribution. In this study, we employed a pairwise method to rank road segments and strike a balance between the accuracy and computational efficiency.

\begin{figure*}
  \centering
  \includegraphics[width=1.00\textwidth]{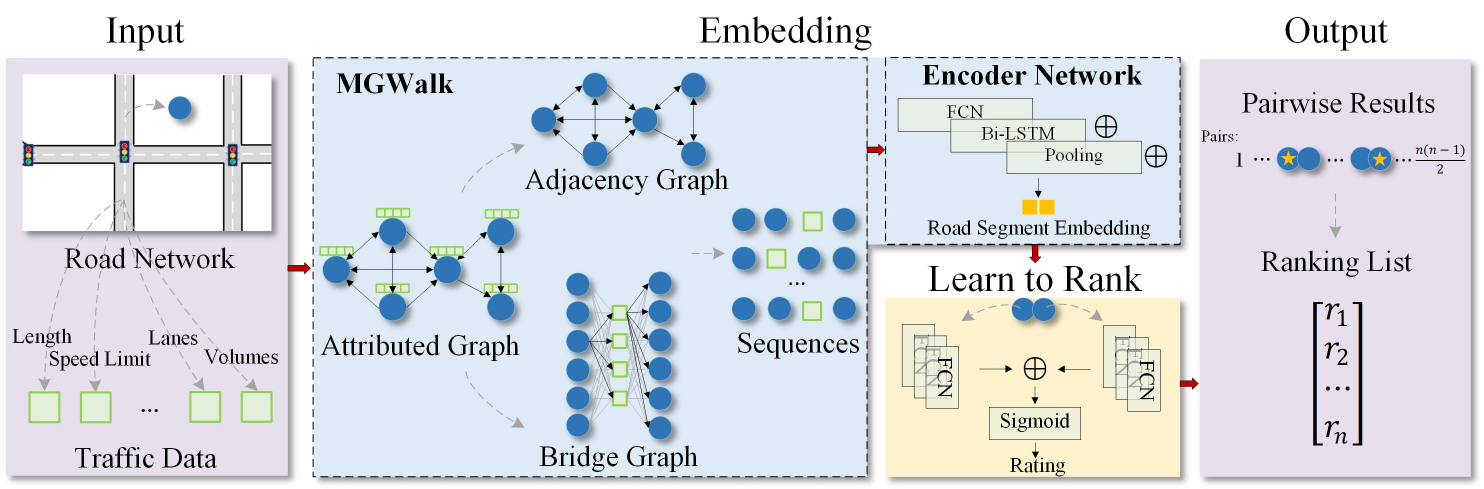}
  \caption{General framework.}\label{fig.1}
\end{figure*}

\subsection{Graph representation learning}\vspace{0.3cm}
Significant progress has been made in the application of graph representation learning for processing complex graphs. For instance, in DeepWalk \cite{perozzi2014deepwalk} and node2vec \cite{grover2016node2vec},  different random walking strategies are utilized to sample nodes, and a skip-gram model \cite{mikolov2013distributed} is applied to learn the node embeddings. In LINE \cite{tang2015line}, the learned vector representation of the nodes preserves both first- and second-order proximity information. HOPE \cite{ou2016asymmetric} adopts an embedding method that preserves high-order proximity to generate graph embeddings in a vector space, while retaining asymmetric transitivity. These methods consider only structural information, and the connections between the nodes are preserved. Studies \cite{mcpherson2001birds, zhang2015influenced} have demonstrated that incorporating the auxiliary information of nodes into graph embeddings can provide valuable insights and improve the performance of graph-level applications. In \cite{hamilton2017inductive}, an inductive representation learning method, namely GraphSAGE, was performed by sampling and aggregating attributes from the local neighborhood of a node. Huang et al. \cite{huang2019graph} proposed an algorithm, namely AttriWalk, for the joint sampling of nodes and attributes and introduced a network structure (GraphRNA) to learn node embeddings. However, AttriWalk tends to sample attributes with higher attribute values and loses relationships between similar nodes. Consequently, the representation capability of this model for nodes is insufficient, rendering it unsuitable for applications in large-scale traffic networks with multiple characteristics. Inspired by this, we consider the impact of attributes and attribute values on node similarity from the perspective of traffic networks and then propose a sampling algorithm to assist in solving the node importance ranking problem.

\section{Methodology}\label{section.3}
The objective of this study is to learn the importance ranking of road segments. We propose a ranking framework (MGL2Rank) that integrates the topology and attribute information of traffic networks. Fig. \ref {fig.1} shows the details of this framework, which comprises an embedding module and a ranking module. The embedding module contains a sampling algorithm and an encoding network, and the ranking module is employed to learn the importance ranking. Before introducing our method, we first define the concepts used in this study and then formally define the problem.

\textbf{Definition 1} (\emph{Road network}): A road network is a directed, unweighted, and attributed graph represented by $ G=(V,E,A) $, where $ V=\{v_1,v_2,...,v_{|V|}\} $ is a set of nodes representing road segments, and $ E $ is a set of directed edges representing the connectivity between these road segments. The adjacency matrix is denoted by $ M\in[0,1]^{|V| \times |V|} $, where $ m_{ij} \in M $ equals 1 if there exists a directed edge from $ v_i $ to $ v_j $, indicating that $ v_i $ can reach $ v_j $; otherwise, it equals 0; $ A \in \mathbb{R}^{|V| \times |U|} $ is a node attribute matrix, where $U$ is a set of attributes, and $ a_{ik} \in A $ is the value of the attribute $ \sigma_k $ corresponding to node $ v_i $. The node attributes comprise road attributes (Definition 2) and traffic attributes (Definition 3).

\textbf{Definition 2} (\emph{Road attributes}): The road attributes are the physical features of a road segment. For a road segment $ v_i \in V $, the road attributes are represented by a vector. In this study, the road attributes include the speed limit, number of lanes, and total length of $ v_i $.

\textbf{Definition 3} (\emph{Traffic attributes}): The traffic attributes of a road network describe the information reflected in traffic volumes during a certain period. The traffic attributes are denoted by a vector. In this study, we consider the total traffic volume and average speed of $ v_i $ during a given period as the traffic attributes.

\textbf{Problem Definition} (\emph{Ranking the importance of nodes}): Given a road network and any pair of road segments $(v_i,v_j)$, we aim to develop a deep learning-based model 
$ 
func_\theta$ with parameter $\theta$, which can obtain a rating that $v_i$ is more important than $v_j$, denoted by $I( r_i > r_j)$, where $r_i$ represents the importance of $v_i$. The model $ 
func_\theta$ can be defined as follows.
\begin{equation}\label{eq1}
  \setlength{\abovedisplayskip}{5pt}
  \setlength{\belowdisplayskip}{5pt}
  \ I( r_i > r_j) := func_\theta (G, (v_i,v_j))
\end{equation}
Finally, we can learn a list of $n$ nodes participating in the ranking in descending importance, denoted by $R$.

\subsection{
MGWalk algorithm}\vspace{0.3cm}
\begin{figure*}[t]
  \centering
  \includegraphics[width=0.95\textwidth]{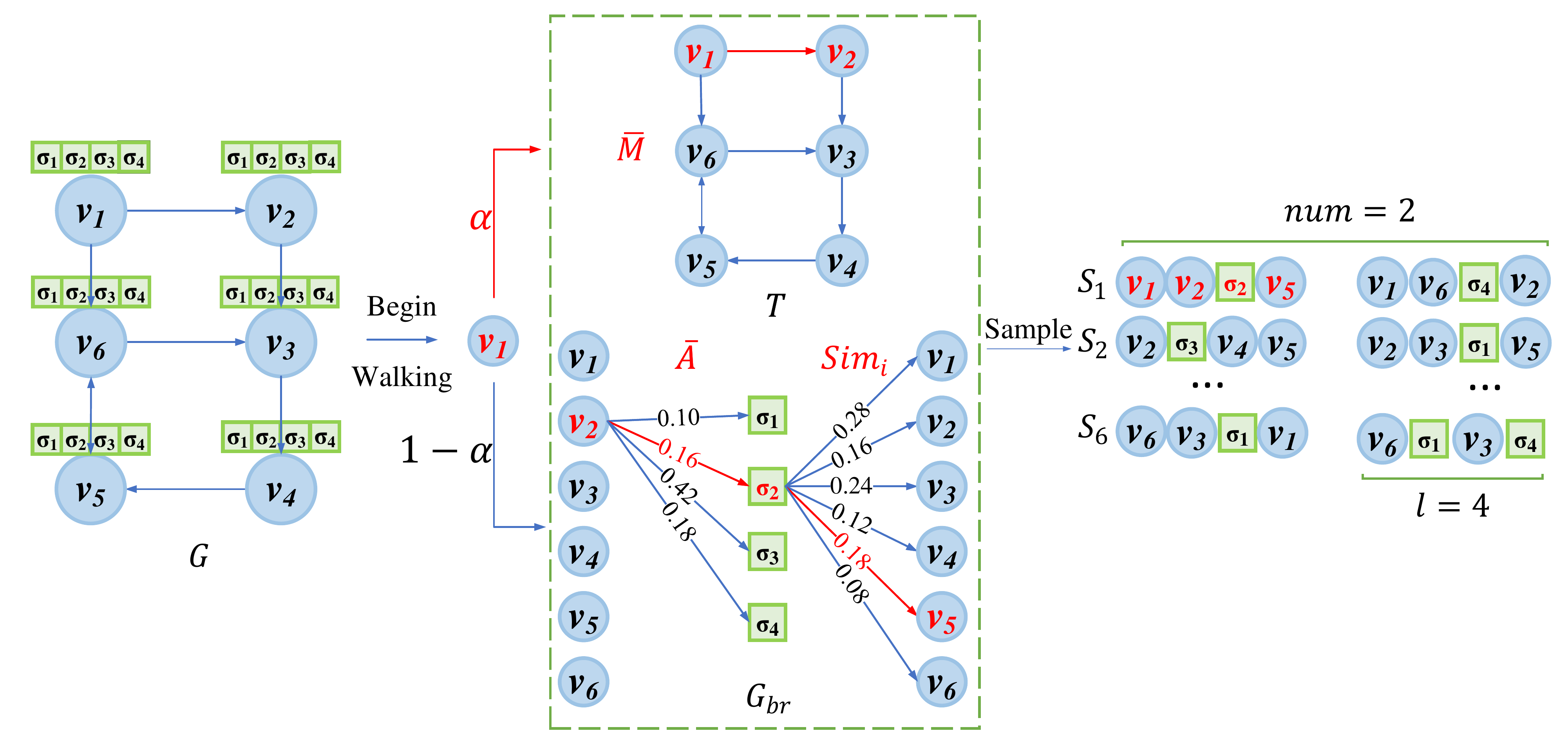}
  \caption{Walking and sampling procedure in MGWalk.\label{Fig.2}}
\end{figure*}
We propose a sampling algorithm, called MGWalk, based on the concept of random walking to establish associations between road segments. This algorithm converts node attributes into nodes on a graph and utilizes the attribute values to guide the sampling. Fig. \ref{Fig.2} shows the walking and sampling procedures of MGWalk. We represent the adjacency relationships of $ G $ as an adjacency graph $ T=(V, E) $ and construct a bridge graph $ G_{br}=(V \cup U, E') $, which is a weighted undirected graph to model the association between nodes based on attributes. $E'$ is a set of edges, the nodes $ v_i \in V $ and $ \sigma_k \in U $ are directly connected by an edge if node $ v_i $ has an attribute $ \sigma_k $. The weights of the edges depend on the attribute matrix $A$.

First, We normalize each column of the transposed adjacency matrix $ M^T $ using the $ l_1 $ norm, denoted by $ \bar{M} $. Each row of the transposed attribute matrix $A^T$ is then normalized using the $l_1$ norm denoted by $ \bar{A}$. The matrix $\bar{A}$ represents the relative value of all the attributes of a node and effectively alleviates the problem of sampling bias toward an attribute with a higher value. The nodes participating in the sampling process include the $|V|$ nodes and $|U|$ attributes of $G$.
The walking starts from node $v_i$, and a walking bias factor $\alpha$ determines the probability of walking based on $T$. MGWalk samples the nodes by iteratively performing the following steps.

\textbf{Walking based on $T$ \textbf{with $\alpha$: }}We take one step from node $v_i$ to node $v_j$, and the sampling probability of this step is defined using the following equation:
\begin{equation}\label{eq2}
  \setlength{\abovedisplayskip}{6pt}
  \setlength{\belowdisplayskip}{6pt}
  \ P(v_i \rightarrow v_j) = \frac { \bar{m}_{ji}}{\sum_{p=1}^{|V|} \bar{m}_{pi}}
\end{equation}
Here, $ \bar{m}_{ji} \in \bar{M} $ is the value in the $ j^{th} $ row and $i^{th}$ column of $ \bar{M} $, indicating the connectivity between $v_j$ and $v_i$. Notably, there are no isolated nodes in the adjacency graph $T$. Therefore, the denominator in Eq. \eqref{eq2} is non-zero. In this way, we can obtain the adjacency relationships between the nodes.

\textbf{ Walking based on $G_{br}$ \textbf{with $ 1-\alpha $:}} We take two steps, and the specific process is as follows.

(1) The first step is taken from node $v_i$ to attribute $\sigma_k$, and the sampling probability in this step can be calculated using the equation:
\begin{equation}\label{eq3}
  \setlength{\abovedisplayskip}{6pt}
  \setlength{\belowdisplayskip}{6pt}
  \ P(v_i \rightarrow \sigma_k ) = \frac { \bar{a}_{ki}}{\sum_{q=1}^{|U|}  \bar{a}_{qi}}
\end{equation}
Here, $\bar{a}_{ki} \in \bar{A} $ is the value in the $k^{th}$ row and $i^{th}$ column of $\bar{A}$. The denominator in Eq. \eqref{eq3} is non-zero because each node has at least one attribute. $P(v_i \rightarrow \sigma_k)$ reflects the relative contribution of each attribute to $v_i$. The higher the contribution of an attribute to $v_i$, the greater the possibility that this attribute will be sampled.

(2) We take the second step from attribute $\sigma_k$ to node $v_j$. We define the absolute difference between the values of the attribute $ \sigma_k $ for node $v_j$ and $v_i$, denoted by $d_{ij}^k$, as represented by the following equation:
\begin{equation}\label{eq4}
  \setlength{\abovedisplayskip}{5pt}
  \setlength{\belowdisplayskip}{5pt}
  \setlength{\abovedisplayskip}{5pt}
  \setlength{\belowdisplayskip}{5pt}
  \ d_{ij}^k = |\bar{a}_{kj}-\bar{a}_{ki}|
\end{equation}
\setlength{\baselineskip}{1.5em}
Here, a lower value of $d_{ij}^k$ indicates a higher similarity between nodes, resulting in a higher probability of sampling node $v_j$. $Sim^i \in \mathbb{R}^{|V| \times |U|} $ is the walking probability matrix from attributes to nodes. The $j^{th}$ row and $k^{th}$ column in $Sim^i$ represents the walking probability from attribute $\sigma_k$ to node $v_j$, denoted by $sim_{kj}^i$, which is calculated using Eq. \eqref{eq5}. By doing so, nodes with similar attributes are sampled.
\begin{equation}\label{eq5}
  \setlength{\abovedisplayskip}{5pt}
  \setlength{\belowdisplayskip}{5pt}
  \ sim_{kj}^i = \begin{cases}
      \frac{1-d_{ij}^k}{\sum_{v=1}^{|V|} d_{iv}^k},& \quad if \;\thinspace\bar{a}_{kj} \neq 0 \\
      \phantom{1-}0,& \quad if \;\thinspace \bar{a}_{kj} = 0
  \end{cases}
\end{equation}
\begin{algorithm}[t]
    \renewcommand{\algorithmicrequire}{\textbf{Input:}}
    \renewcommand{\algorithmicensure}{\textbf{Output:}}
	\caption{The MGWalk Algorithm.}
	\label{alg:1}
	\begin{algorithmic}[1]
		\REQUIRE The adjacency matrix $M$, attribute matrix $A$, walking bias factor $\alpha$, number of sequences $num$, and sequences length $L$
		\ENSURE The set of sampled sequences $S=\{S_1,S_2,\ldots,S_{|V|}\}$
		\STATE $\bar{M}$, $\bar{A}$ $\leftarrow$ $l_1$ norm to normalize each column of $M^T$ and each row of $A^T$
		\STATE 	Calculate the sampling probability matrix $Sim^i$ from the attributes to nodes using Eqs. \eqref{eq4} and \eqref{eq5}
		\STATE Calculate the sampling probability matrix $P$ using Eq. \eqref{eq6} 
	   \STATE qList=\{\} and JList=\{\} 
/* in the alias algorithm, qList is a probability table, JList is an alias table */
        \STATE $ S=\{\} $
		\FOR{$v_i \in V$} 
        \STATE $ S_i=\{\} $
        \FOR{$j=1:{num}$}
        \STATE $ s=\{\} $
        \STATE Append AliasMethod $(L,P)$ to $s$ /* obtain a sequence of $L$ nodes by $P$ */
        \STATE Append $s$ to $S_i$
        \ENDFOR
        \STATE Append $S_i$ to $S$
        \ENDFOR
        \STATE Return $S=\{S_1,S_2,\ldots,S_{|V|}\}$
  \end{algorithmic}
\end{algorithm}
In summary, the sampling probability matrix of MGWalk can be expressed using Eq. \eqref{eq6}, where each column of $P$ is normalized. When $\alpha$ is 1, the walking mechanism is a random walk. Otherwise, the similarity between the nodes is considered in the walking mechanism. Finally, we utilize the aliasing method \cite{devroye1986sample} to sample from a discrete probability distribution.
\begin{equation}\label{eq6}
  \setlength{\abovedisplayskip}{5pt}
  \setlength{\belowdisplayskip}{5pt}
  \setlength{\abovedisplayskip}{5pt}
  \setlength{\belowdisplayskip}{5pt}
  \ P= \begin{bmatrix}
      \alpha\bar{M}& (1-\alpha)Sim^i\\
      (1-\alpha)\bar{A}& 0
  \end{bmatrix} \in \mathbb{R}_+^{(|V|+|U|)\times(|V|+|U|)}
\end{equation}

We introduce the sampling settings in MGWalk. For node $v_i$, we represent a set of the sampled sequences as $S_i$, which comprises $num$ sequences of length $L$. The implementation process is presented in Algorithm \ref{alg:1}, according to the walking mechanism described above.

\subsection{
Embedding network}
\begin{figure}[t]
  \centering
  \includegraphics[width=\linewidth]{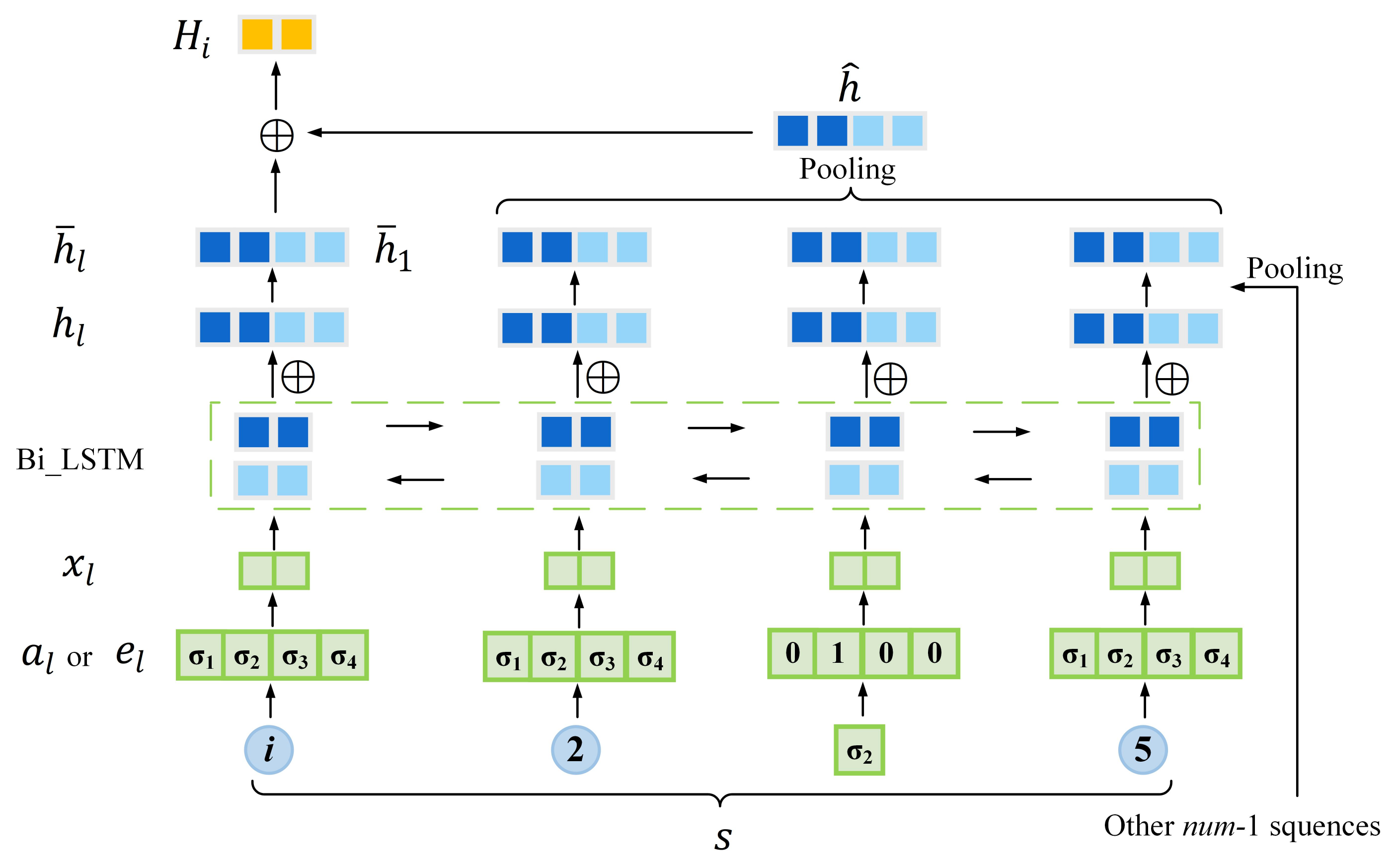}
  \caption{Architecture of the representation learning module.\label{Fig.3}}
\end{figure}

\vspace{0.2cm}
We introduce a representation learning module that uses sampled sequences to obtain node embeddings. One of the essential components of this module is the bidirectional LSTM network. The bidirectional nature of the network enables it to simultaneously  capture forward and backward contextual information from sampled sequences, which is consistent with the interdependencies between a road segment and its upstream and downstream segments. Consequently, the bidirectional LSTM can enhance the representation of the hub role of the nodes in the entire road network. Fig. \ref{Fig.3} shows the architecture of this module.

MGL2Rank learns the embedding of node $v_i$ on $S_i$. For a sequence $s=(s_1,s_2,\ldots,s_L)$ in $S_i$, each element in $s$ is first mapped to a node attribute vector or a one-hot vector representing the attribute. Specifically, if $s_l \in s$ corresponds to a road segment node, we map it to its attribute vector, denoted by $a_l$; if $s_l \in s$ represents an attribute, we encode it as a one-hot vector denoted by $e_l$. Subsequently, we obtain the initial vector representation of $s_l$, denoted by $x_l$, which is formulated as follows:
\begin{equation}\label{eq7}
  \ x_l = \begin{cases}
      Tanh(a_l W_{l x}+b_{lx}), \quad if \; s_l \in V \\ Tanh(e_l W_{l x}+b_{lx}), \quad if \; s_l \in U
  \end{cases}
\end{equation}
Here, $ W_{lx} $ and $b_{lx}$ are the weight matrix and bias of the layer, respectively.  Thus, we can obtain the initial vector representation of each node in $s$, denoted by $x=(x_1,x_2,\ldots,x_L)$.

We employ the bidirectional LSTM to learn the contextual information of $x$ by calculating the forward hidden state sequence ${\mathop{x}\limits ^{\rightarrow}} $ and backward hidden state sequence $ {\mathop{x}\limits ^{\leftarrow}}$. The forward hidden state vector $ {\mathop{h}\limits ^{\rightarrow}{_l}} $ is calculated using \eqref{eq8}-\eqref{eq12}:
\begin{equation} \label{eq8}
  \setlength{\abovedisplayskip}{8pt}
  \setlength{\belowdisplayskip}{8pt}
  \ i_l = \sigma ( {x_l}W_{xi} + \mathop{h} \limits ^{\rightarrow} 
  {_{l-1}}W_{hi} + b_i)
\end{equation}
\begin{equation}\label{eq9}
  \setlength{\abovedisplayskip}{8pt}
  \setlength{\belowdisplayskip}{8pt}
  \ f_l = \sigma ( {x_l}W_{xf} + {\mathop{h}\limits ^{\rightarrow}} {_{l-1}}W_{hf} + b_f)
\end{equation}
\begin{equation}\label{eq10}
  \setlength{\abovedisplayskip}{8pt}
  \setlength{\belowdisplayskip}{8pt}
  \ c_l = f_l \odot c_{l-1} + i_l Tanh( {x_l}W_{xc} + {\mathop{h}\limits ^{\rightarrow}} {_{l-1}}W_{hc} + b_c)
\end{equation}
\begin{equation}\label{eq11}
  \setlength{\abovedisplayskip}{8pt}
  \setlength{\belowdisplayskip}{8pt}
  \ o_l = \sigma ( {x_l}W_{xo} + {\mathop{h}\limits ^{\rightarrow}} {_{l-1}}W_{ho} + b_o)
\end{equation}
\begin{equation}\label{eq12}
  \setlength{\abovedisplayskip}{8pt}
  \setlength{\belowdisplayskip}{8pt}
  \ {\mathop{h}\limits ^{\rightarrow}{_l}} = o_l \odot Tanh(c_l)
\end{equation}
Here, $\sigma$ is the sigmoid function. $i_l$, $f_l$, $c_l$ and $o_l$ are the the output results for input gate,
forget gate, internal state, and output gate. $W_{xi}$, $W_{xf}$, $W_{xc}$, and $W_{xo}$ are weight matrixes of each gate. $W_{hi}$, $W_{hf}$, $W_{hc}$, and $W_{ho}$ are  the recurrent weights. $b_i$, $b_f$, $b_c$, and $b_o$ are the biases of each gate. The operation "$\odot$"  represents the element-wise product. The backward hidden state sequence $ {\mathop{h}\limits ^{\leftarrow}{_l}} $ is calculated using the same equations. The output of the bidirectional LSTM is obtained using a concatenation operation.
\begin{equation}\label{eq13}
  \setlength{\abovedisplayskip}{8pt}
  \setlength{\belowdisplayskip}{8pt}
  \ h_l =  {\mathop{h}\limits ^{\rightarrow}} {_l} \oplus {\mathop{h}\limits ^{\leftarrow}} {_l}
\end{equation}
Thus, we can obtain $num$ sequences of hidden state vectors with length $L$, denoted by a set $\{\eta_1,\eta_2,\ldots, \noindent  \eta_{num}\}$. We perform average pooling for each dimension to obtain $(\bar{h}_1,\bar{h}_2,\ldots,\bar{h}_L) $. Next, $(\bar{h}_2,\ldots,\bar{h}_L)$ is further merged into a vector $\hat{h}$ by average pooling. The final embedding of node $v_i$, denoted by $H_i$, is obtained by concatenating operation:
\begin{equation}\label{eq14}
   \setlength{\abovedisplayskip}{8pt}
   \setlength{\belowdisplayskip}{8pt}
   \ H_i = \bar{h}_1 \oplus \hat{h}
\end{equation} 
Here, $H_i \in \mathbb{R}^{hdim}$, where $hdim$ represents the final embedding dimension, contains not only the adjacency of the node $v_i$ but also the associations of attributes with other nodes in the graph.

\subsection{Ranking module}\vspace{0.3cm}
We introduce a pairwise ranking module that utilizes the obtained embeddings to solve this importance ranking problem. Fig. \ref{Fig.4} shows the architecture of the ranking module. The specific details are as follows:
\begin{figure}[t]
  \centering
  \includegraphics[width=1.00\linewidth]{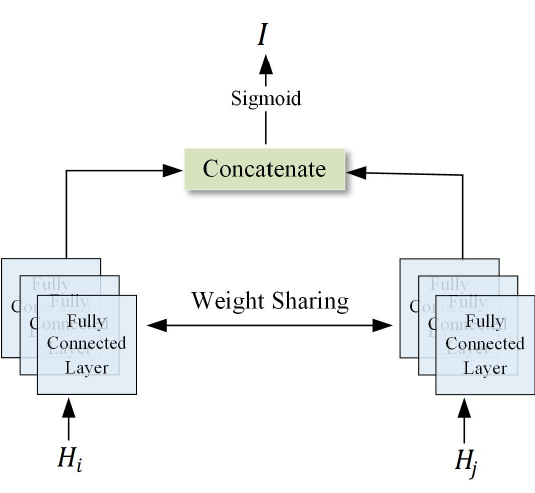}
  \caption{Architecture of the ranking module.\label{Fig.4}}
\end{figure}

A Siamese neural network is used to determine the relative importance of the nodes. A Siamese neural network comprises two subnetworks $g_{\theta^{'}}$ with shared parameters $ \theta^{'} $, each containing three fully connected layers. We first input the embeddings of two nodes, $H_i$ and $H_j$, into the ranking module. $g_{\theta^{'}}(H_i)$ and $g_{\theta^{'}}(H_j)$ are then concatenated into a global feature vector, as we aim to capture the relationships and differences between the node features. Finally, the output of the ranking module can be expressed as:
\begin{equation}\label{eq15}
   \ I( r_i > r_j) = sigmoid(g_{\theta^{'}}(H_i) \oplus g_{\theta^{'}}(H_j))
\end{equation}
here $I$ indicates the rating that $v_i$ is more important than $v_j$. 

The objective of model training is to minimize the error between the predicted ranking and the ground-truth. To achieve this, we define label $\bar{I}_u$ to represent the actual ranking for the $u^{th}$ pair of nodes as follows:
\begin{equation}\label{eq17}
   \bar{I}_u = \begin{cases}
      \ 1,\quad if \; scr_u^1 > scr_u^2 \\
      \ 0,\quad if \; scr_u^1 \le scr_u^2
  \end{cases}
\end{equation}
Here, $scr_u^1$ and $scr_u^2$ represent the actual importance scores of the $u^{th}$ pair of nodes.  We utilize the binary cross entropy as the loss function, denoted by $\mathbf{L}_f$, as follows:
\begin{equation}\label{eq18}
    \mathbf{L}_f=- \frac{1}{n^{'}} \sum_{u=1}^{n{'}}(\bar{I}_u \log I_u + (1-\bar{I}_u) \log (1-I_u)
\end{equation}
Here, $I_u$ represents the predicted ranking for the $u^{th}$ pair of nodes, $ n^{'} $ is the number of node pairs participating in the ranking.

\section{Experiments}\label{section.4}\vspace{0.3cm}
In this section, we describe the experiments conducted in our study to evaluate the proposed framework. We discuss our synthetic dataset in Subsection \ref{subsection 4.1}, followed by a description of the experimental design in Subsection \ref{subsection 4.2}. In Subsection \ref{subsection 4.3}, we show and analyze the results of comparative experiments. The case study is detailed in Subsection \ref{subsection 4.4}. An ablation study is introduced in Subsection \ref{subsection 4.5} to analyze the contribution of each component, and a sensitivity analysis is conducted in Subsection \ref{subsection 4.6}. The methods were implemented in Python 3.8 and executed on a server with an Intel Xeon(R) Platinum 8370 CPU and an RTX 3090 24G GPU.

\subsection{Dataset}\vspace{0.3cm} \label{subsection 4.1}
\begin{figure}[t]
  \centering
  \includegraphics[width=1.0\linewidth]{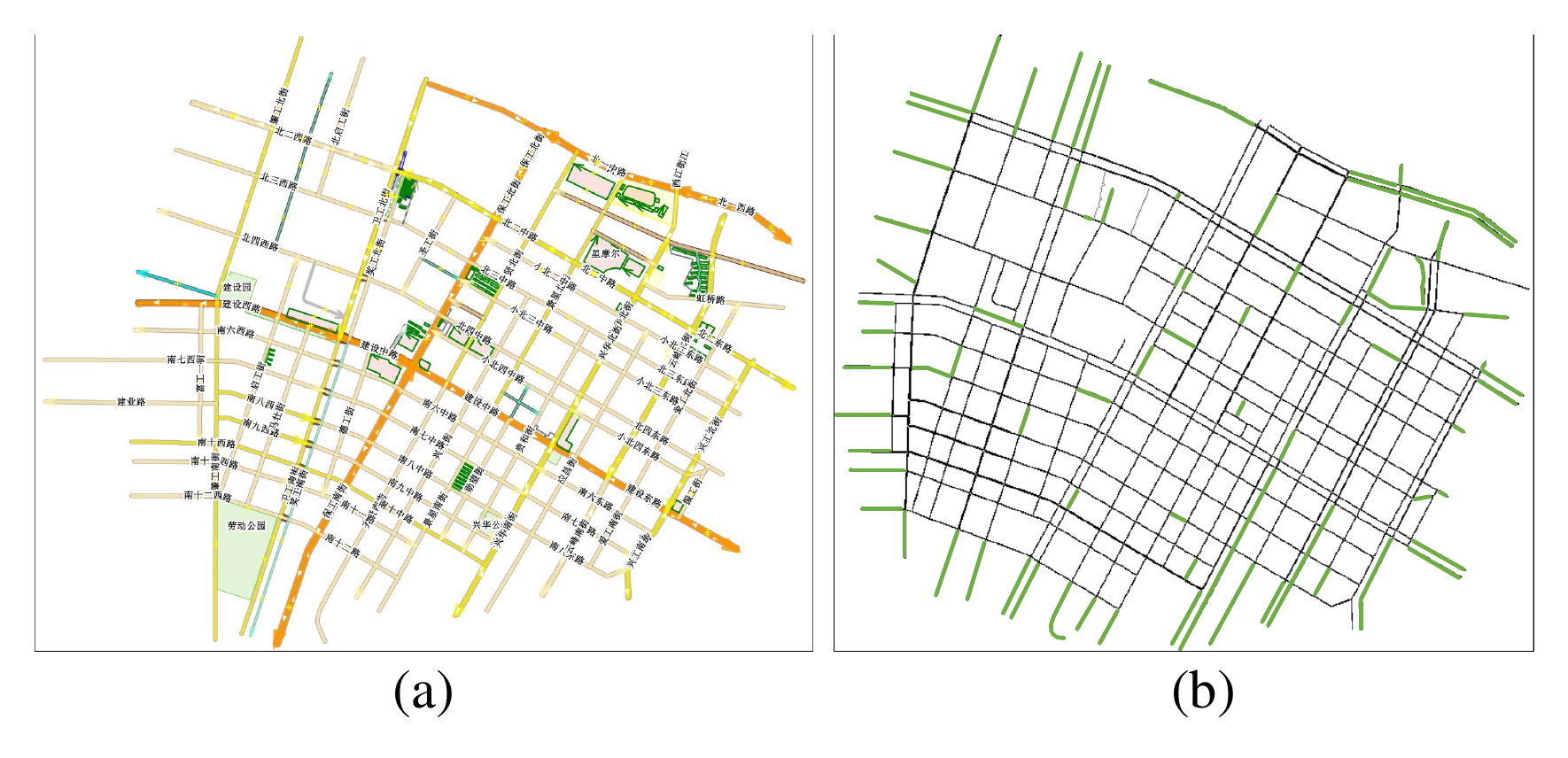}
  \caption{Real-world and simulated road network of Shenyang city.\label{Fig.5}}
\end{figure}

We constructed a synthetic dataset based on the real-world road network of Tiexi District in Shenyang City using the traffic simulator SUMO \cite{lopez2018microscopic}. The road network comprises 1004 road segments and 377 intersections, as shown in Fig. \ref{Fig.5}. Inspired by origin-destination (OD) flow inference model, we incorporated both points-of-interest (POIs) and OD information into our dataset to simulate the traffic distribution. Road segments located near certain POIs, such as subway stations, schools, and shopping malls, were selected as origins or destinations, which are indicated by green lines in Fig. \ref{Fig.5}(b). Next, we simulated a 1 h period during peak hours and introduced 7200 cars into the simulated road network based on the OD information. Road segments without trajectory data were excluded from the dataset. Table \ref{tab:1} provides the statistics of the dataset.

\begin{table}[t]
  \centering
  \caption{Statistics of the dataset.}
  \label{tab:1}
  \begin{tabular}{ccccccc}
    \hline
    dataset & $\left| \rm V \right|$ & $\left| \rm E \right|$ & $\left| \rm U \right|$ & origin & destination & cars\\
    \hline
    Network\_SY & 929 & 3168 & 16 & 95 & 99 & 7200 \\
    \hline
\end{tabular}
\end{table} 

\begin{figure}[t]
  \centering
  \includegraphics[width=0.98\linewidth]{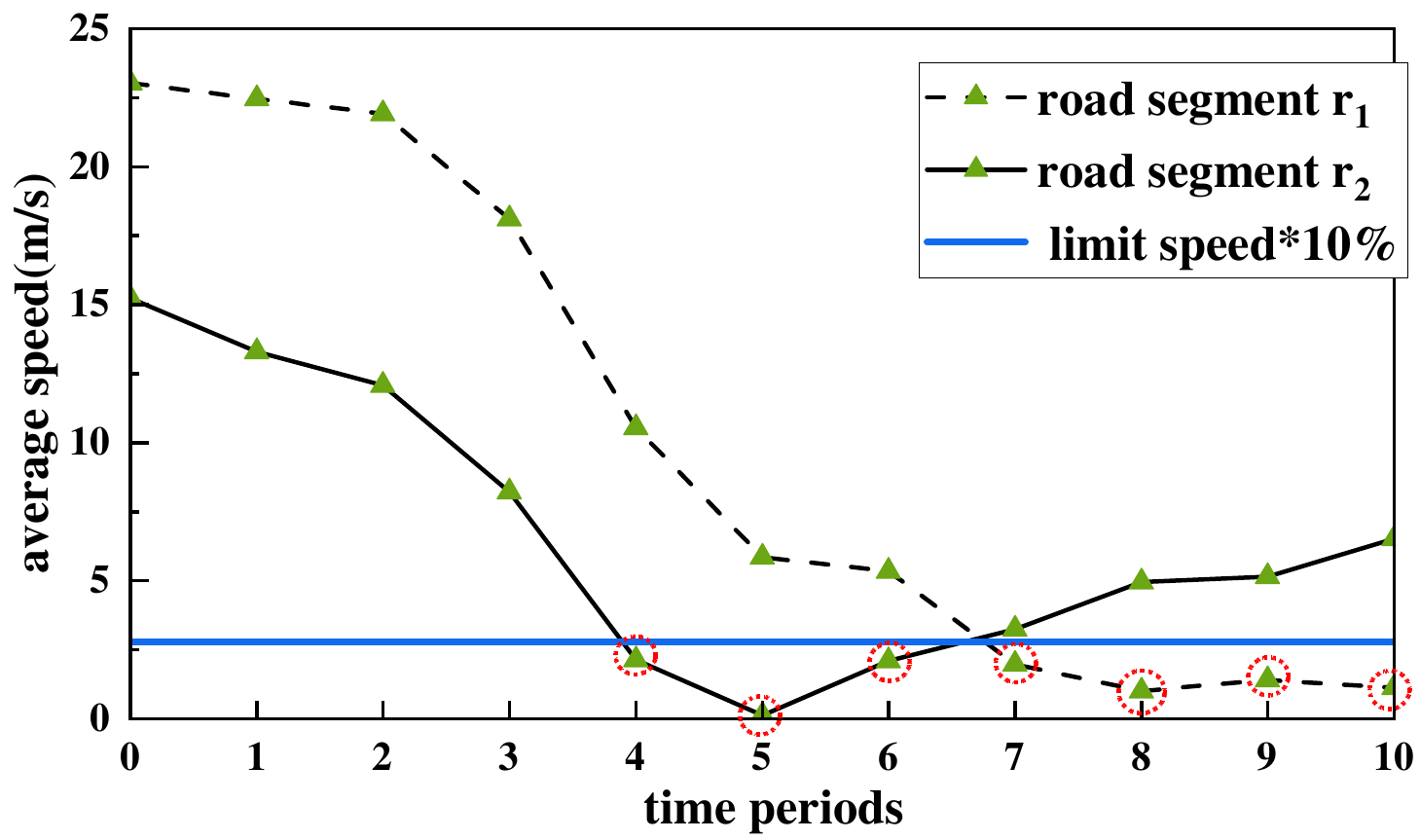}
  \caption{Average speed of road segments over time.\label{Fig.6}}
\end{figure}
To obtain the ground-truth ranking of the node importance for all the road segments, we simulated the failures of each road segment by reducing its traffic capacity to $10\%$. We then measured the impact of these failures on the overall traffic efficiency of a road network over a specific period. If the average speed of a road segment fell below $10\%$ of its speed limit, it was considered a failure. Fig. \ref{Fig.6} illustrates the average speed fluctuations in the two additional road segments during different periods following the failure of a specific road segment. The road segment $v_i$ failed during periods $t$ = 4, 5, and 6, whereas the road segment $v_j$ failed during periods $t$ = 7, 8, 9, and 10. Finally, the importance score of the road segment $v_i$ is defined as follows:
\begin{equation}\label{eq20}
   \ aff_{i}=\sum_{t=1}^{Q} \gamma ^t \bar{n}_t
\end{equation}
Here, $\gamma$ is a decay factor, $\bar{n}_t$ is the number of failed road segments in the entire road network during the period $t$, $Q$ is the total number of periods. Setting $\gamma$ to 0.9 means that the contribution of each period to the importance score decreased by $10\%$. In this way, we quantified both the congestion level and capacity of the congestion propagation for each road segment.

\subsection{Experimental design}\vspace{0.3cm} \label{subsection 4.2}
We conducted two groups of comparative experiments and an ablation study. First, we compared MGL2Rank with the following baseline methods in terms of the ranking performance:
\begin{itemize}
\item Degree centrality (DC) \cite{freeman2002centrality}: It measures the importance of a node by calculating the number of edges connected to it. Nodes with higher degree centrality values are generally considered more important in a network.
\item Betweenness centrality (BC) \cite{freeman1977set}: It evaluates the importance of a node based on the frequency of its occurrence in the shortest network paths.
\item PageRank \cite{Page1999ThePC}: PageRank iteratively calculates the importance of a node based on the number and importance of the other linked nodes.
\item TraNode2vec  
\cite{huang2022traffic}: TraNode2vec evaluates the importance of intersections based on clustering and representation learning.
\item RankNet \cite{burges2005learning}: RankNet is a deep learning-based method that predicts the order of a set of objects by training on pairwise comparisons.
\end{itemize}

To verify the effectiveness of our embedding module for the ranking task, we compared it with the following baseline methods:
\begin{itemize}
\item DeepWalk \cite{perozzi2014deepwalk}: DeepWalk obtains node sequences by performing truncated random walks on a graph and applies a skip-gram model to learn the node representation.
\item GraphRNA \cite{huang2019graph}: It is an embedding framework that obtains sampling sequences by walking between nodes and attributes based on their shared attribute.
\end{itemize}

The ranking performance was evaluated using two standard metrics: micro-F1 and macro-F1. We additionally utilized $Diff(R)$ \cite{xu2018discovery} as an evaluation metric, which was proposed in our previous study. $Diff(R)$ can be defined as follows,
\begin{equation}\label{eq19}
    Diff(R) = \frac{\sum_{\tau=1}^{n}| Z(v_\tau,R)-Z(v_\tau,desc(R))|}
   {\lfloor n^2/2 \rfloor}
\end{equation}
where $v_\tau$ is the $\tau^{th}$ node in the predicted ranking list $R$; $desc(R)$ is a list of nodes in descending order of the importance, and 
$Z(v_\tau,R)$ is the position index of $v_\tau$ in $R$. The ranking performance improved as $Diff(R)$ decreased.

Stratified random sampling was employed for all the nodes to ensure a balanced distribution of the node importance across all the datasets. We allocated $70\%$ of the nodes as the training set, $15\%$ as the validation set, and the remaining $15\%$ as the test set. The embedding dimension $hdim$ was set to 8, learning rate to 0.001, dropout rate to 0.45, batch size to 64, and number of epochs to 100. For MGWalk, we set walking bias factor $\alpha$ to 0.0001, the number of sampled sequences $num$ to 150, and the sequence length $L$ to 4.

\subsection{
Results}\vspace{0.3cm} \label{subsection 4.3}
Table \ref{tab:2} presents the ranking performances of the different methods in terms of three metrics. First, we compared MGL2Rank with the ranking baseline methods. Traditional methods exhibited a poor performance because they are not learnable. Among deep learning-based methods, RankNet exhibited the best performance. Compared with RankNet, MGL2Rank showed improvements of $8.96\%$ and $5.93\%$ in terms of Micro-F1 and $Diff$, respectively. This improvement can be attributed to the learnable embedding module in MGL2Rank, which is absent in RankNet. Notably, MGL2Rank outperformed the baseline methods in terms of the overall ranking distribution, as evidenced by $Diff$. We then compared our embedding module with baseline embedding methods. MGL2Rank showed improvements of $22.20\%$ and $ 8.62\% $ in terms of Micro-F1, and $45.81\%$ and $ 2.29\%$ in terms of $Diff$, respectively.
This indicates that our embedding module can effectively capture valuable information regarding the structural relationships and attribute associations of road segments.
\begin{table}[t]
  \centering
  \caption{Ranking performance of different methods.}
  \label{tab:2}
  \begin{tabular}{ccccc}
    \hline
    \multicolumn{2}{c}{\multirow{2}{*}{Methods}} & \multicolumn{3}{c}{Metric}\\
    \cmidrule(lr){3-5}
    && Micro-F1 & Macro-F1 & Diff \\
    \hline
    \rule{0pt}{8pt}
    \multirow{4}{*}{Ranking} & DC & 0.4510 & 0.4428 & 0.4973\\ 
    & BC & 0.5110 & 0.5058 & 0.4257 \\
    & PageRank & 0.5453 & 0.5313 & 0.4658\\
    & TraNode2vec & 0.7158 & 0.7126 & 0.3203\\
    & RankNet & 0.7192 & 0.7161 & 0.3145\\
    \hline
    \rule{0pt}{10pt}
    \multirow{2}{*}{Embedding} & DeepWalk & 0.6146 & 0.6094 & 0.4329\\ 
    & GraphRNA & 0.7219 & 0.7140 & 0.3037\\
    \hline 
    \rule{0pt}{10pt}
    \multirow{3}{*}{Ablation} & NoMG & 0.7168 & 0.7124 & 0.3184\\ 
    & NoBiLSTM & 0.7682 & 0.7678 & 0.3164\\
    & NoEmb & 0.7269 & 0.7231 & 0.3034\\
    \hline 
    \rule{0pt}{10pt}
    \multirow{1}{*}{} & MGL2Rank & $\mathbf{0.7900}$ & $\mathbf{0.7866}$ & $\mathbf{0.2969}$\\ 
  \bottomrule
\end{tabular}
\end{table}

\begin{figure*}[t]
  \centering
  \includegraphics[width=1.00\textwidth]{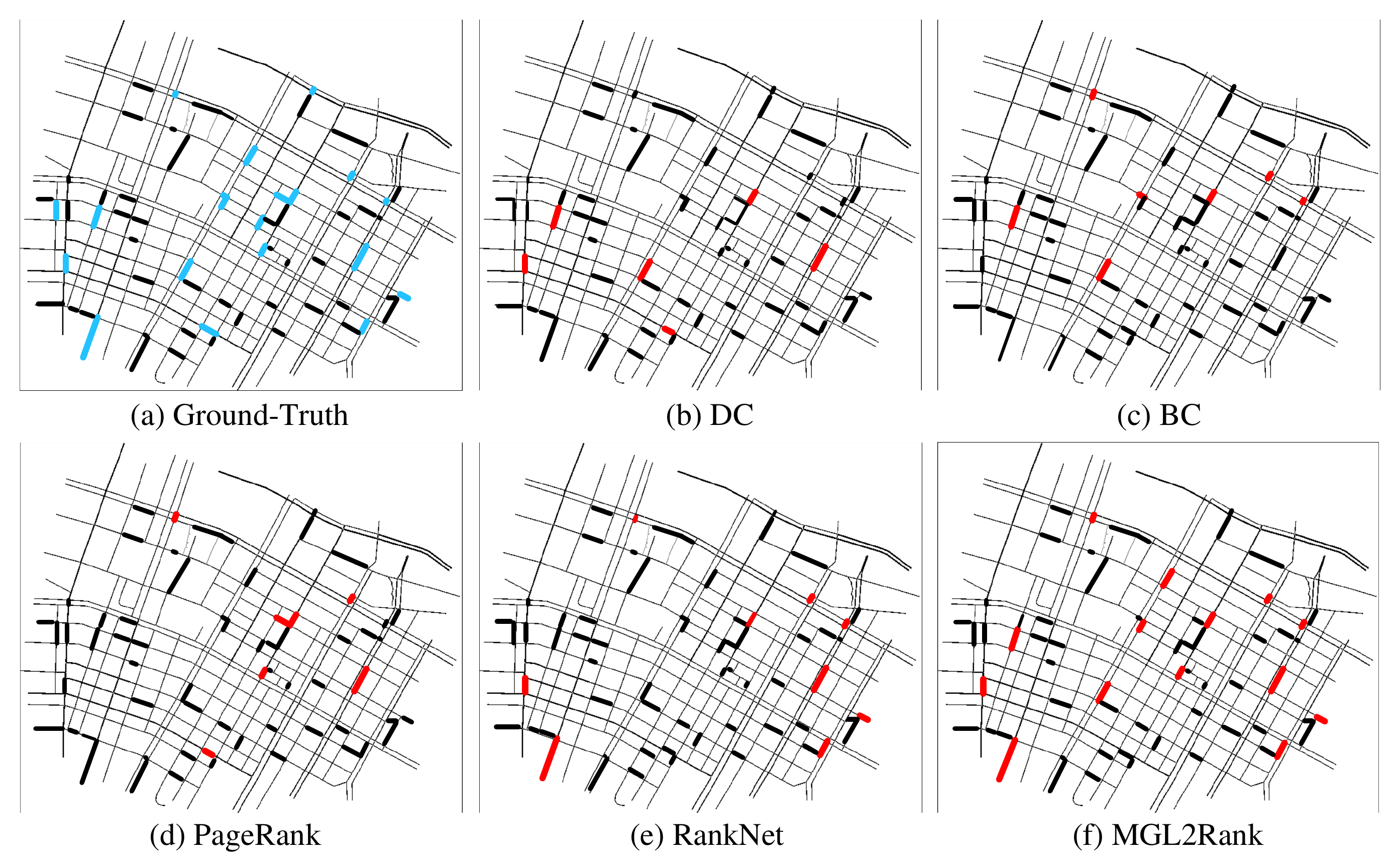}
  \caption{Actual top-20 important road segments and ranking results of different methods. (All bolded lines indicate the road segments participating in the ranking.)\label{Fig.7}}
\end{figure*}
\vspace{-0.4cm}We selected a test set containing 64 road segments and visualized the ranking results of the different methods, as shown in Fig. \ref{Fig.7}. The actual top-20 important road segments are highlighted with bold blue lines in Fig. \ref{Fig.7}(a). The bold red lines in Figs. \ref{Fig.7}(b)-(f) indicate the top-20 important road segments that can be correctly identified by the five methods (DC, BC, PageRank, RankNet, and MGL2Rank). The bold red lines indicate more accurate ranking results. Our observations are as follows: (1) The ranking performances of degree centrality, betweenness centrality, and PageRank are relatively poor. This means that centrality-based metrics may not necessarily be high for actual important road segments. (2) RankNet can identify more top-20 important road segments compared with the traditional methods. However, its limited feature mining capability decreases its performance. (3) Among all the methods, MGL2Rank could identify the highest number of important road segments, with 14 out of the top-20 being accurately identified. This suggests that the embedding method based on multi-graph fusion can effectively capture location-based dependencies and attribute associations between road segments, thereby yielding a better ranking performance.
\begin{figure*}[t]
  \centering
  \includegraphics[width=1.00\textwidth]{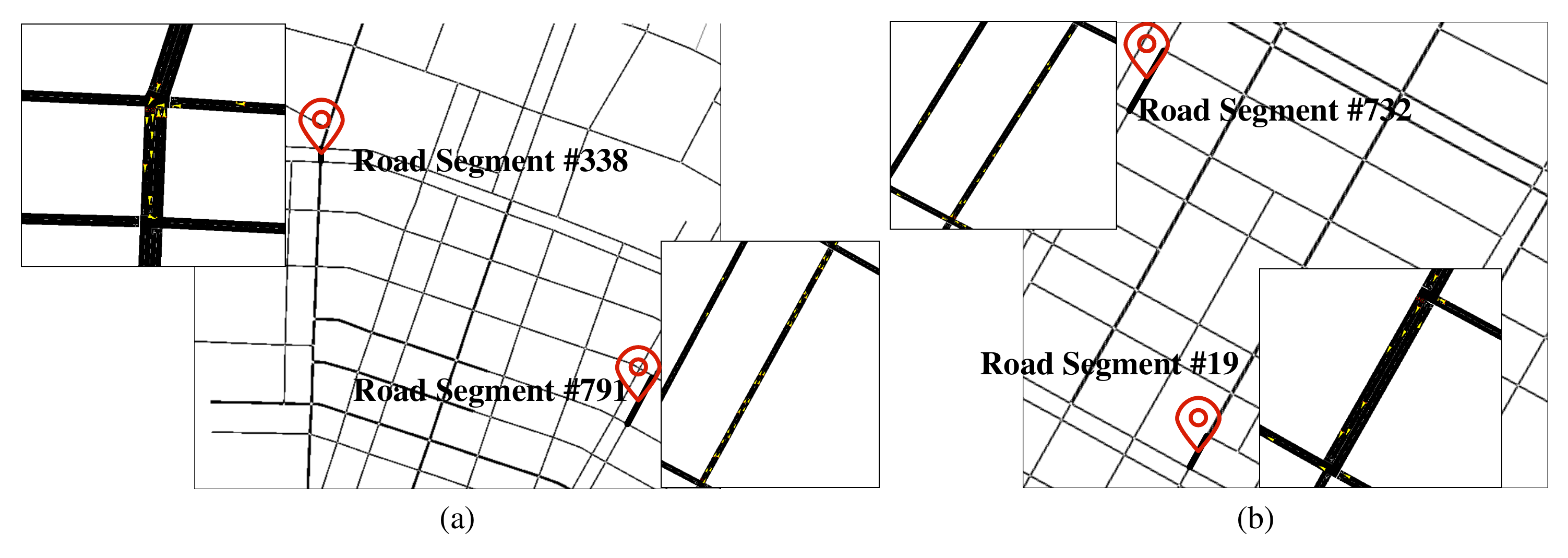}
  \caption{Topologies and attributes of selected road segments.\label{Fig.8}}
\end{figure*}
\subsection{Case study}\vspace{0.3cm}\label{subsection 4.4}
We selected two pairs of road segments ranked by the MGL2Rank framework for the analysis, as shown in Fig. \ref{Fig.8}. In Fig. \ref{Fig.8}(a), although Node $\#791$ has a lower traffic volume, it is more important than Node $\#338$. This can be attributed to the fact that Node $\#791$ is connected to the upstream Node $\#192$ with more lanes and higher traffic volumes. The failure of Node $\#791$ led to congestion at Node $\#192$, subsequently affecting more road segments. As shown in Fig. \ref{Fig.8}(b), Node $\#19$ with a lower traffic volume is more important than Node $\#732$. This is because Node $\#19$ is in an area with numerous short road segments, leading to rapid congestion propagation. These observations reveal that the importance of a road segment is reflected in both its traffic capacity and congestion propagation speed. The proposed method considers the aforementioned factors from the following two perspectives: (1) The traffic capacity of a road segment is mainly influenced by its own attributes, such as the number of lanes and traffic volumes. Thus, we incorporated diverse traffic characteristics into the embedding by multi-graph fusion to enhance the representation of the traffic capacity. (2) The congestion propagation speed of a road segment is influenced by the topology and associations between road segments based on their attributes. Road segments with similar attributes exhibited similar congestion propagation speeds. In this way, MGL2Rank can accurately identify nodes that play an important role in the overall transportation system.
\begin{figure*}[t]
  \centering
  \includegraphics[width=0.65\textwidth]{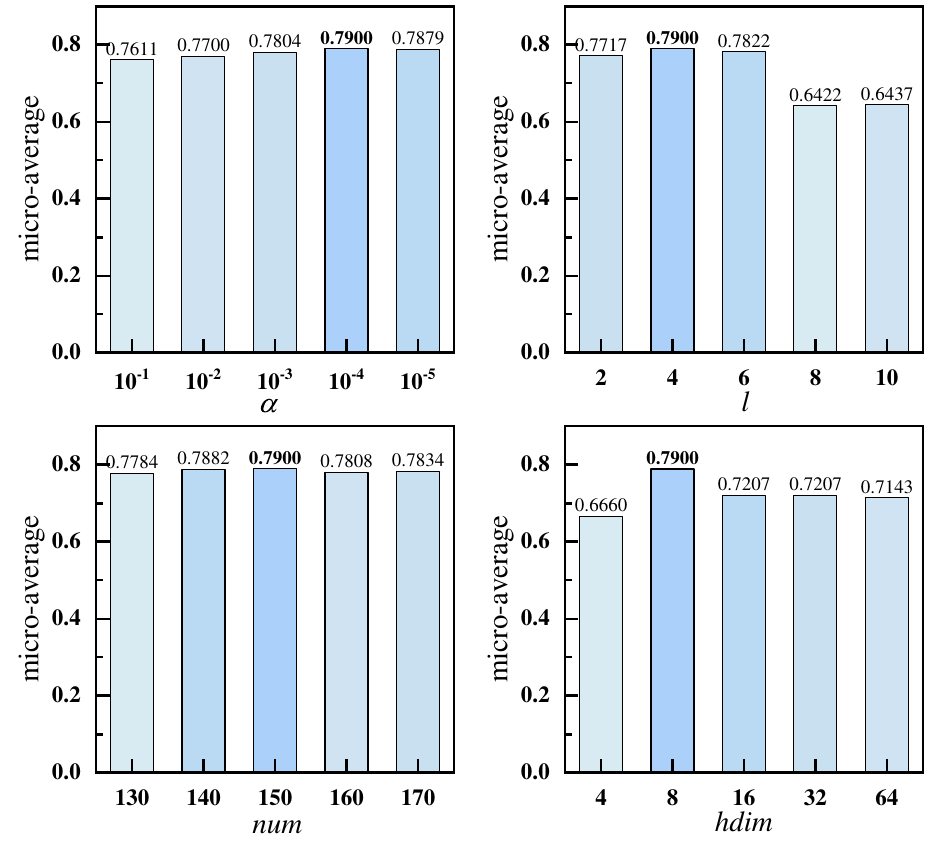}
  \caption{Performance under different hyperparameters.\label{Fig.9}}
\end{figure*}
\subsection{Ablation study}\label{subsection 4.5}
We used three variants of MGL2Rank to evaluate the contributions of each component. Table \ref{tab:2} summarizes the results of the ablation study.
\begin{itemize}
\item  NoMG: This variant replaces MGWalk with a truncated random walk for sampling while keeping the other modules the same as those in MGL2Rank.
\item NoBiLSTM: In this variant, there is no bidirectional LSTM network in the embedding module; the other modules are the same as those in MGL2Rank.
\item  NoEmb: This variant retains only the ranking module, which learns to rank based on the initial features of road segments.
\end{itemize}

As shown in Table \ref{tab:2}, we make the following observations by comparison with MGL2Rank: (1) NoMG exhibited a $9.27\%$ decrease in Micro-F1, indicating that MGWalk has an advantage in extracting and integrating the attributes of similar road segments. (2) NoBiLSTM exhibited a $2.76\%$ decrease in Micro-F1, indicating that the bidirectional LSTM in the embedding module can learn contextual information for node associations. MGWalk contributes more to the overall performance. (3) Compared with MGL2Rank, NoMG exhibited a $9.27\%$ decrease in Micro-F1 because the random walk in NoMG lost meaningful node associations. The collaboration between MGWalk and bidirectional LSTM played a crucial role in the performance of MGL2Rank.

\subsection{Sensitivity analysis}\vspace{0.3cm} \label{subsection 4.6}
Sensitivity studies were performed on the necessary parameters, including the walking bias factor ($\alpha$), number of sampled sequences ($num$), sequence length ($L$), and embedding dimension ($hdim$). Fig. \ref{Fig.9} shows the performance under different hyperparameter settings. The following observations can be made from this figure: (1) The model exhibited the best performance when $\alpha$ is 0.0001. The performance decreased slightly when $\alpha$ deviated from this value. An appropriate value of $\alpha$ can help balance the topological information and attributes. In Particular, if the number of attributes is significantly lower than the number of nodes, the value of $\alpha$ should be higher. (2) When $L$ is 4, MGL2Rank exhibited the best performance. The performance declined when $L$ decreased to 2, as meaningful information regarding the node associations became limited. When $L$ exceeded 6, the performance began to decline owing to the introduction of interfering information. (3) The performance remained stable as $num$ increased from 130 to 170. The performance was optimum when $num$ was set to 150. (4) When the embedding dimension was set to 8, the model exhibited the highest ranking accuracy. The performance started to decrease when $hdim$ exceeded 8, indicating that higher $hdim$ values do not necessarily improve the performance.
 
\section{Conclusion and future work}\vspace{0.3cm} \label{section.5}
We developed a deep learning-based framework (MGL2Rank) to rank the importance of nodes in urban road networks. This framework comprises a sampling algorithm (MGWalk) that captures the topology of the road network and the complex associations between road segments using multi-graph fusion. Next, we introduced an efficient encoder network to learn the embeddings of the road segments. A ranking module was proposed to learn the importance ranking of the road segments based on the obtained embeddings. Finally, we constructed a dataset based on the road network of Shenyang City. The results of experiments conducted on this dataset demonstrated that MGL2Rank outperformed typical baseline methods. In the future, we plan to explore road networks at different scales to further evaluate our framework. We also aim to investigate the potential of our method to address other transportation-related problems such as traffic congestion prediction and route planning optimization.

\bibliographystyle{IEEEtran}

\end{document}